# FWLBP: A Scale Invariant Descriptor for Texture Classification

Swalpa Kumar Roy[†], *Student Member, IEEE*, Nilavra Bhattacharya, *Student Member, IEEE*,
Bhabatosh Chanda, Bidyut B. Chaudhuri, *Life Fellow, IEEE*, Dipak Kumar Ghosh, *Member, IEEE*

**Abstract**—In this paper we propose a novel texture descriptor called Fractal Weighted Local Binary Pattern (FWLBP). The fractal dimension (FD) measure is relatively invariant to scale-changes, and presents a good correlation with human viewpoint of surface roughness. We have utilized this property to construct a scale-invariant descriptor. Here, the input image is sampled using an augmented form of the local binary pattern (LBP) over three different radii, and then used an indexing operation to assign FD weights to the collected samples. The final histogram of the descriptor has its features calculated using LBP, and its weights computed from the FD image. The proposed descriptor is scale invariant, and is also robust in rotation or reflection, and partially tolerant to noise and illumination changes. In addition, the local fractal dimension is relatively insensitive to the *bi-Lipschitz* transformations, whereas its extension is adequate to precisely discriminate the fundamental of texture primitives. Experiment results carried out on standard texture databases show that the proposed descriptor achieved better classification rates compared to the state-of-the-art descriptors.

✦

## 1 INTRODUCTION

CONSTRUCTING effective texture feature is a major challenges in computer vision [1]. It has received a lot of attention in the past decades due to its value in understanding how human beings recognize textures. A crucial issue of texture analysis is constructing an effective texture representation. There are primarily five methodologies of texture representation: statistical, geometrical, structural, model-based, and signal processing based [2]–[4]. In recent literature, two categories of texture classification approaches dominate current research, namely texture descriptor based methods [5], [6] and deep learning based methods. In the second category, a convolutional neural network (CNN) is trained to classify the texture images [7], [8]. The deep learning based methods offer good classification performance, however it has following limitations: it requires a large amount of data and hence is computationally expensive to trains. Basically the complex models take weeks to train using several machines equipped with expensive GPUs. Also, at present, no strong theoretical foundation on topology/training method/flavor /hyper-parameters for deep learning exist in the literature. On the other hand, the descriptor based methods have the advantage of data independence, ease to use, and robustness to real-life challenges such as illumination and scale changes. In this category of methods the design of effective texture descriptor is regarded as extremely important for texture recognition.

A picture may be captured under geometric and photometric varying conditions. An ideal model for representing and classifying textures should be capable to capture fundamental perceptual textures' properties. It should be robust enough against changes, like view-point alteration, luminance variation, image rotation, reflection, scale change, and geometry of the underlying surface. Attention has been focused on designing of local texture descriptors to accomplish local invariance [9]–[12]. The search for invariant descriptors started in the 1990s [13]. Kashyap and Khotanzad first proposed a circular autoregressive dense model [14] for rotation invariant texture classification. Many other models including multi-resolution [15], hidden Markov model [16], and Gaussian Markov model [17] were explored to study rotation invariance. More recently Varma and Zisserman [11] proposed a texton based method for rotation invariant texture classification where the texton dictionary is learned from a set of filter responses of training samples, and then texture image is classified based on its texton distribution. They also introduced another texton based method [12] where local image patch is used to represent the feature directly. In addition some works have been done on scale and affine invariant feature extraction in texture classification. Among them Varma and Garg [18] extracted a local fractal vector for each pixel, and computed a statistical histogram. Liu and Fieguth [19] applied random projection for densely sampled image patches, and extracted the histogram signature. Yao and Sun [20] normalized statistical edge feature distribution to resist the variation in scale. Lazebinik *et al.* [21] and Zhang *et al.* [9] detected Harris and Laplacian key points for extracting texture signatures. Recently, global scale invariant feature extraction methods drew more attention because local scale normalization is slow because of pixel by pixel operations are needed there. Xu *et al.* [22] and Quan *et al.* [6] have classified the image pixels into multiple point sets by gray intensities or local feature descriptors. Instead of extracting scale invariant features, pyramid histograms with shifted matching scheme was proposed by some researchers [10], [23]. Roy *et al.* [24] introduced a complete dual-cross pattern (CDCP) to address the scale and rotational effects in unconstrained texture classification. Generally, a texture can be characterized by geometric multi-scale self-similar macro-structure (like fractal dimension and LBP). Recently Roy *et al.* [25] proposed Local morphological pattern (LMP), a combination of basic mathematical morphology operations (i.e. opening and closing) and the computation of the LBP over the resulting images for texture classification. Geometric structure of object in the image vary with scale changes while illumination change does not change the object structure. Several studies have been done using fractal based texture classification [18], [26]–[28] but the drawback of these techniques are that the value of fractal dimension is continuous, therefore a quantization stage is required to compute the histogram. However, an accurate quantization depends upon a excessive training with huge number of training samples.

*S. K. Roy and B. B. Chaudhuri are with the Optical Character Recognition Laboratory, Computer Vision and Pattern Recognition Unit at Indian Statistical Institute, Kolkata 700108, India (email: swalpa@ieee.org; bbc@isical.ac.in).*

*N. Bhattacharya is with the School of Information, University of Texas at Austin, TX 78712, USA (email: nilavra@ieee.org).*

*B. Chanda is with the Image Processing Laboratory, Electronics and Communication Sciences Unit at Indian Statistical Institute, Kolkata 700108, India (email: chanda@isical.ac.in).*

*D. K. Ghosh is with the Department of Electronics and Communication Engineering at National Institute of Technology Rourkela, Rourkela 769008, India (email: dipak@ieee.org).*

[†] *Corresponding author, (email: swalpa@ieee.org).*

In this work, we approach the quantization problem by proposing a new descriptor called Fractal Weighted Local Binary Pattern (FWLBP) based on a commonly used method of combining fractal dimension proposed by Chaudhuri and Sarkar [29]. This method uses differential box-counting (DBC) to measure the texture surface instead of directly. In DBC the texture surface measures at different scale by means the number of counted boxes of different size, which can cover the whole surface.

The contributions of this work are summarized below:

- We propose a simple, effective, yet robust fractal-dimension based texture descriptor called Fractal Weighted Local Binary Pattern (Sec. 3.1) for texture classification.
- The FWLBP descriptor achieves sufficient invariance to address the challenges of scale, translation, and rotation (or reflection); and is also partially tolerant to noise and illumination for texture classification.
- The proposed algorithm takes a reasonable amount of time in the feature extraction stage (Sec. 3.2).
- We experimented and observed that our proposed descriptor outperforms the traditional LBP based and other state-of-the-art methods on different texture databases, such as KTH-TIPS and CUReT (Sec. 4).

The rest of the paper is organized as follows. Sec. 2 deals with the mathematical background of fractal dimension, and our approach to convert the input image to its FD form. The proposed FWLBP feature extraction scheme is presented in Sec. 3. In Sec. 4 the performance of texture classification is compared with the state-of-the-art methods. Finally, the conclusions are drawn in Sec. 5.

## 2 FRACTAL DIMENSION

A fractal is a geometrical set whose Hausdorff - Besicovitch dimension is strictly greater than its topological dimension [32]. Mandelbrot introduced the term *fractal* to describe non-Euclidean, self-similar structures. The Fractal Dimension (FD) of a structure gives an idea of its texture complexity, which can be used to measure, analyse and classify shape and texture. Fractal surfaces show the property of self-similarity.

In the Euclidian $n$-space, a set $S$ is said to be self-similar, if $S$ is the union of $N_r(s)$ distinct (non-overlapping) copies of itself, each of which is scaled down by a ratio $r > 0$. Let $N_r(s) = k(\frac{1}{r})^{D(s)}$ (i.e. an exponential function of $r$), where $D(s)$ is a density function, and $k$ is a constant. Then

$$\log N_r(s) = \log k + D(s) \cdot \log \frac{1}{r} \quad (1)$$

Then, the local Fractal Dimension (FD) of $s$ becomes

$$D(s) = \lim_{r \to 0} \frac{\log(N_r(s))}{\log(1/r)}. \quad (2)$$

It is difficult to compute the local Fractal dimension $D$ using Eqn. (2) directly. In our implementation (Fig. 1), we transform the texture images into FD images using the DBC algorithm [30]. To achieve scale-invariance, we construct a Gaussian scale space [33] using Eqn. (3):

$$\Gamma_{x,y,r} = \langle G_r(x,y) | I(x,y) \rangle = \int_{x,y \in \mathbb{R}} G_r(x,y) I(x,y) dx dy. \quad (3)$$

In Eqn (3) a convolution formalism may be used when convolution across the entire image is required. However, here an inner product formalism ($\langle . | . \rangle$) is more convenient where $G_r$ is a Gaussian smoothing kernel with variance $r$:

$$G_r(x) = \frac{1}{r\sqrt{2\pi}} \exp^{\frac{-\|x^2\|}{2\sigma^2}}$$

The convolution operation of the input image with 2**D** Gaussian kernel can be efficiently computed using two passes of the 1**D** Gaussian kernel in the vertical and horizontal directions as 2**D** Gaussian function is separable [34], [35]:

$$G_r(x,y,r) = G_r(x) G_r(y).$$

The scaling operation to obtain $I_r = G_r * I$ (where $*$ represents convolution operator) can be computed efficiently and stably because of the properly localization of the Gaussian both in space and frequency [36] even if the input image (I) is the result of physical measurement, so called directly sampled [37]. After constructing the Gaussian scale space, we use the DBC algorithm on each layer of the scale space to generate a set of intermediate images. We combine the intermediate images into the final FD image using a linear-regression technique adapted from [31].

The input texture image $A_{x,y}(X \times Y)$ is used to generate a Gaussian scale space $\Gamma_{x,y,r}(X \times Y \times L)$ having $L$ layers, with scaling factor $r$ varying from $r_{\min}$, to $r_{\max}$, and total number of layers $L = r_{\max} - r_{\min} + 1$. Each layer is treated as a $R^3$ space, with $(x,y)$ representing the $R^2$ position, and the third coordinate, $g$, describing the gray level. The DBC algorithm is implemented by applying a varying-size non-linear kernel $b_{i,j}(m \times n)$, which operates on the image surface and finds the difference between the maximum ($g_{\max}$) and the minimum ($g_{\min}$) gray values of the region enclosed within the kernel (Fig. 2). Non-negative integers $\alpha$ and $\beta$ are used to center the kernel $b(i,j)$ on the pixel $g_{x,y}$ in the image layer. The kernel is computed as

$$b(i,j) = \sum_{i=-\alpha}^{\alpha} \sum_{j=-\beta}^{\beta} \text{floor}\left[\frac{g_{\max} - g_{\min}}{r}\right] + 1 \quad (4)$$

where $r$ is the scaling factor of the layer $l$, and

$$\alpha = \text{ceil}\left(\frac{m-1}{2}\right), \quad \text{and} \quad \beta = \text{ceil}\left(\frac{n-1}{2}\right). \quad (5)$$

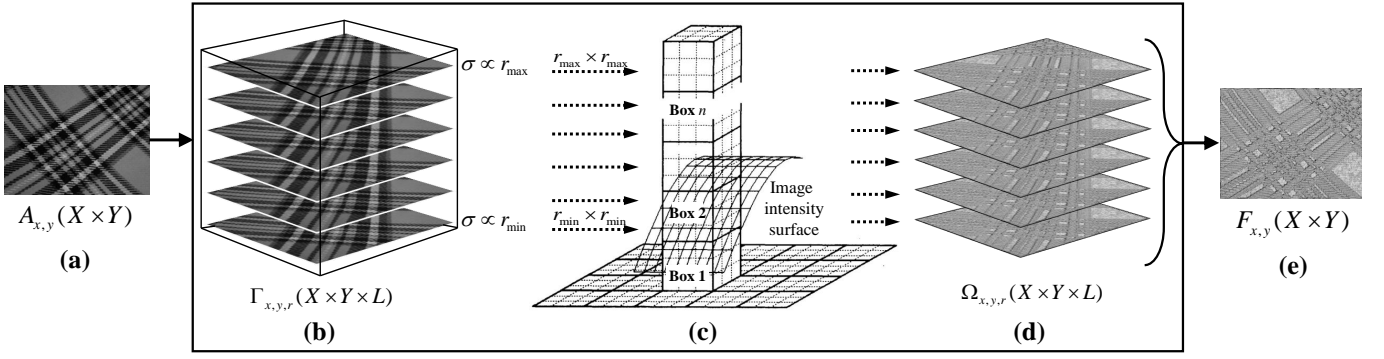

Figure 1: Illustrates the Differential Box Counting (DBC) technique [30] for computing Fractal Dimension. (a) Input texture image. (b) Gaussian Scale Space with six layers, generated from the input image with box sizes from $r_{\min} \times r_{\min}, \ldots, r_{\max} \times r_{\max}$. (c) DBC is used to calculate the FD for each image layer of the scale space. (d) Intermediate images. (e) Final FD image obtained using technique adapted from [31].

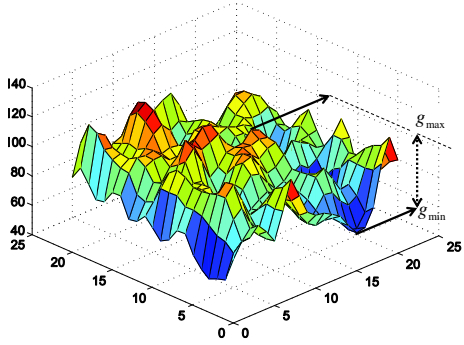

Figure 2: Image intensity surface enclosed within a box, showing the difference between maximum and minimum gray values.

The kernel is applied as

$$\Omega(x,y,r) = \sum_{i=-\alpha}^{\alpha} \sum_{j=-\beta}^{\beta} b(i,j) \Gamma(x+\alpha, y+\beta, r) \left(\frac{L}{r}\right)^2 \quad (6)$$

where $\Omega_{x,y,r}(X \times Y \times L)$ is a matrix of intermediate images, such that the first layer denotes the original image filtered by the kernel with scale $r = r_{\min}$, and the highest layer denoting the image filtered by the kernel with scale $r = r_{\max}$. Thus, we define

$$\Omega(x,y,r) = \begin{bmatrix} g_{11r} & g_{12r} & \cdots & g_{1Yr} \\ g_{21r} & g_{22r} & \cdots & g_{2Yr} \\ \vdots & \vdots & \ddots & \vdots \\ g_{X1r} & g_{X2r} & \cdots & g_{XYr} \end{bmatrix} \quad (7)$$

The slope $\lambda_{x,y}$ of the least square linear regression line between $\Omega(x,y,r)$ and $r$ will denote the pixel value $F(x,y)$ of the FD image $F_{x,y}(X \times Y)$. For this purpose, a column-vector $\Theta$ is defined, such that the first gray values of all the layers in $\Omega(x,y,r)$ comprise vector $\theta_1$, all the second gray values comprise vector $\theta_2$ etc. as shown in Eqn. (8).

$$\Theta = \begin{bmatrix} \theta_1 \\ \theta_2 \\ \vdots \\ \theta_{X \times Y} \end{bmatrix} = \begin{bmatrix} g_{111} & g_{112} & \cdots & g_{11L} \\ g_{121} & g_{122} & \cdots & g_{12L} \\ \vdots & \vdots & \ddots & \vdots \\ g_{XY1} & g_{XY2} & \cdots & g_{XYL} \end{bmatrix} \quad (8)$$

The slope can be determined by computing the sums of squares:

$$\Psi_1 = \sum r^2 - \frac{(\sum r)^2}{L} \quad (9)$$

$$\Psi_2 = \sum r\Theta - \frac{(\sum r)^2 (\sum \Theta)^2}{L}. \quad (10)$$

Finally, the FD image $F_{x,y}(X \times Y)$ is calculated as:

$$F(x,y) = \lambda_{xy} = \sum_{x=1}^{X} \sum_{y=1}^{Y} \frac{\Psi_2}{\Psi_1} \quad (11)$$

## 2.1 Effects of Fractal Transform

An interesting characteristic of the Fractal Transform is its invariance under *bi-Lipschitz transform*. A transform $t: R^2 \rightarrow R^2$ is called a bi-Lipschitz transform, if two constants $0 < \varsigma_1 \leq \varsigma_2 \leq \infty$ are exist such that for any two points $p_1, p_2 \in R^2$,

$$\varsigma_1 \|\varsigma_1 - \varsigma_2\| < \|t(\varsigma_1) - t(\varsigma_2)\| < \varsigma_2 \|p_1 - p_2\| \quad (12)$$

where $\|p_1 - p_2\|$ represent Euclidean metric between $p_1$ and $p_2$. So, any traditional transform (like projective transformation, rotation, translation, texture warping of a regular surface, change in viewpoint, and non-rigid deformation) is a *bi-Lipschitz* transform [38], [39]. Up-sampling or down-sampling by a factor of $n$ are also special cases of the *bi-Lipschitz* transform, where $\varsigma_1 = \varsigma_2 = n$. Then, we can drive the following theorem:

**Theorem 1.** *Let us consider an image as $f(p), p \in R^2$. Let the fractal dimension and length of point $p$ be $D$ and $L$, respectively. For a sampling function $t(p) = np$, with $n > 1$ represents up-sampling, and $n < 1$ represents down-sampling, the fractal dimension and length at point $t(p)$ become*

$$D_t = D, \quad L_t = L + D \cdot \log n \quad (13)$$

**Algorithm 1:** Algorithm to compute FD image of input texture image.

**Data:** Input texture image, $A_{x,y}(X \times Y)$
**Result:** Matrix $F_{x,y}(X \times Y)$ containing FD values of each pixel of $A$
1  initialize ($r_{min} = 2, r_{max} = 7, L = r_{max} - r_{min} + 1$);
2  **while** ($r \geq r_{min}$ and $r \leq r_{max}$) **do**
3  $\quad G_r$ = Gaussian_Kernel$([r, r], r/2)$;
4  $\quad \Gamma_{(x,y,r)} = \langle G_r(x,y) | I(x,y) \rangle$;
5  $\quad$ /* Update DBC kernel $b$ */
6  $\quad b(i,j) = \sum\limits_{i=-\alpha}^{\alpha} \sum\limits_{j=-\beta}^{\beta} \text{floor}\left[\frac{g_{max} - g_{min}}{r}\right] + 1$;
7  $\quad$ /* Compute $\Omega(x,y,r)$ Matrix of intermediate FD values, where $r$ represents the scaling factor on layer $l$ and $l \in [1, L]$ */
8  $\quad \Omega(x,y,r) = \sum\limits_{i=-\alpha}^{\alpha} \sum\limits_{j=-\beta}^{\beta} b(i,j)\Gamma(x+\alpha, y+\beta, r)\left(\frac{L}{r}\right)^2$;
9  **end while**
10  Convert the 3-$\mathbb{D}$ matrix $\Omega_{x,y,r}(X \times Y \times L)$ to a column vector $\Theta$, as per Eqn. (8).;
11  /* Compute sum of square */
12  $\Psi_1 = \sum r^2 - \frac{(\sum r)^2}{L}$;
13  $\Psi_2 = \sum r\Theta - \frac{(\sum r)^2 (\sum \Theta)^2}{L}$;
14  Compute FD image $F(x,y) = \sum\limits_{x=1}^{X} \sum\limits_{y=1}^{Y} \frac{\Psi_2}{\Psi_1}$;

where $L = \log k$.

*Proof.* For a point $p$, we have $\log N_r(p) = D \cdot \log(\frac{1}{r}) + L$ (from Eqn. (1)). By the definition of $t(p)$ we know that $t$ is invertible and $\|t(p_1) - t(p_2)\| = n \|p_1 - p_2\|$. The bi-Lipschitz constant ensures that $N_r(t(p)) = N_{\frac{r}{n}}(p)$. Hence we can write,

$$\begin{aligned}
\log N_{\frac{r}{n}}(p) &= D \cdot \log \frac{n}{r} + L \\
\Rightarrow \log N_r(t(p)) &= D \cdot \log \frac{1}{r} + D \log n + L \\
\Rightarrow \log N_r(t(p)) &= D_t \cdot \log \frac{1}{r} + L_t \\
\Rightarrow D_t &= D, \quad L_t = L + D \cdot \log n
\end{aligned} \quad (14)$$

□

Hence, from the above theorem, the fractal dimension of an image is invariant to scale changes, or in other words, invariant to local *bi-Lipschitz* transforms, but not the fractal length of the image. Theoretically, the fractal dimension of an image remains unchanged when it is sampled-up with a factor of $n$ and vice-versa. This is our main motivation to take advantage of the fractal dimension in our proposed descriptor.

## 3 PROPOSED TEXTURE DESCRIPTOR

Scale change can make an impressive impact on texture appearance. There is no locally invariant texture descriptor that can deals scale changes of such image intensity surface. The goal of this paper is to build texture descriptor which is robust to scale changes in the intensity surface of natural texture. Even local fractal dimension is a powerful measurement of surface "roughness" of a natural image, though the fractal dimension alone is not sufficient to describe natural texture. The FD yields continuous values and so a quantization is needed. The quantization process can be done by computing the feature distribution from all the training samples to know about the distribution of feature space. Finally, some threshold value is calculated to divided the feature space into fixed number of bins and the quantization of test FD image is done according to the training threshold values. Fig. 3 illustrates quantization process of the feature space using four bins. There are the following important limitations due to the quantization process, which need to be addressed carefully.

- To identify the threshold values for each bin, it require a pre-training stage.
- Texture images are captured under different geometric and photometric varying conditions. So, the quantization process fully depends on the training samples.
- The choice of number of bins is a challenging task although there are some techniques [40] to select it. It is always a trade-off between optimal number of bins for the best of accuracy and feature size.

Since the distribution has a finite number of entries, choosing a few bins may fail to provide sufficient discriminative power while large number of bins may lead to sparse and unreliable results. Hence, feature matching becomes computationally inefficient. The method is explained in the following sections.

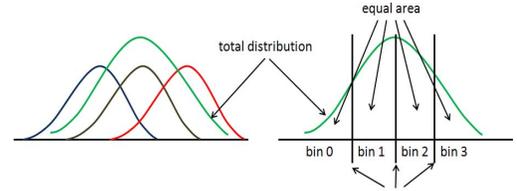

Figure 3: Quantization procedure of the feature space

### 3.1 Fractal Weighted Local Binary Pattern (FWLBP)

A good texture descriptor should have four powerful characteristics: reliability, independence, discrimination, and small size. Hence, we propose a simple, efficient, yet robust texture-descriptor called the Fractal Weighted Local Binary Pattern (FWLBP) and offer a solution of aforementioned problem using the concept of local image patterns exhibiting self-similarity. The discrimination power of the proposed descriptor is enhanced by using the fractal dimensions as weights of the histogram where the sampling method of indexing is an augmentation of the native LBP [40]. We generated multi-scale indexing images at different scales by varying the sampling radius ($R \in [R_{min}, R_{max}]$). However, to keep reasonable complexity, we have kept number of sampling points ($N$) constant across all the scales. Since fractal dimension is a logarithmic function (Eqn. (2)), it helps to reduce effects of illumination, because the $\log$ function expands the values of darker pixels and compresses the brighter pixels in the image. As a result, the FD values are spread more uniformly [41]. Thus, the proposed descriptor is insensitive to scale, translation, rotation or reflection, illumination; and is also noise-tolerant.

After generating the scale-invariant FD image (Fig. 4b) of the original texture using the DBC technique (Sec. 2),

we proceed to compute the proposed descriptor – Fractal Weighted Local Binary Pattern (FWLBP). We generate multi-

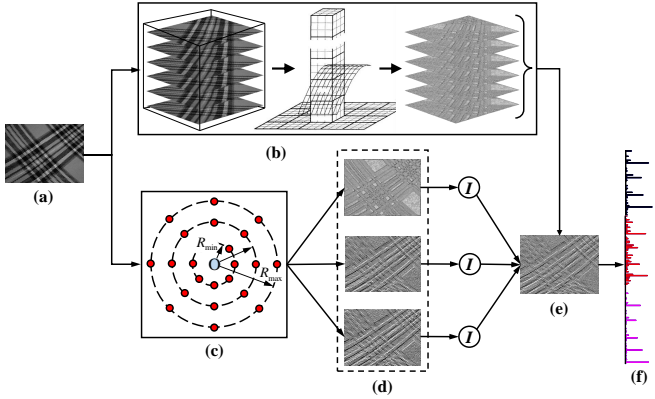

Figure 4: Proposed FWLBP algorithm. (a) input texture image (b) computation of FD image using Gaussian Scale Space and DBC algorithm (c) taking $\text{LBP}^{R,N}$ samples from input image using $R \in [R_{\min}, R_{\max}]$ and constant $N$ (d) LBP images for varying $R$ (e) indexing output FD image with the different LBP images to calculate fractal weights (f) final feature vector

resolution LBP images (Fig. 4(d)) from the input texture image by varying the sampling radius $R \in [R_{\min}, R_{\max}]$ (Fig. 4(c)). The LBP for a given pixel at location $(x, y)$ in the input image $A_{x,y}(X \times Y)$ is computed by comparing its gray value $a_{x,y}$ with a set of $N$ local circularly and equally spaced neighbours $P_{x,y}^{R,N}$ placed at a radius $R$ ( $R \in [R_{\min}, R_{\max}]$) around the pixel. LBP is computed for only those pixels whose all $N$ local neighbours lie within the image, and not for pixels along the image boundary. The final LBP value of a given pixel is calculated as

$$\text{LBP}_{x,y}^{R,N} = \sum_{n=1}^{N} 2^{n-1} \times sign(P_{x,y}^{R,N} - a_{x,y}) \quad (15)$$

where *sign* is a *unit step* function to denote whether a given input is positive or not, and defined as

$$sign(\xi) = \begin{cases} 1, & \xi \geq 0 \\ 0, & \xi < 0 \end{cases}.$$

The range of $\text{LBP}_{x,y}^{R,N}$ depends on the number of neighbouring sampling points ($N$) around the pixel $(x, y)$ at radius $R$ to form the pattern, and its value lies in between 0 to $2^{N-1}$. In other words, the range of LBP is $[0, 2^{N-1}]$. Fig. 4(d) shows the computed local binary patterns (i.e. $\text{LBP}_{x,y}^{R,N} | R = [R_{\min}, R_{\max}]$) for a candidate texture image (Fig. 4(a)).

Finally, we use an *indexing operation* $\mathcal{I}$ to combine each of the LBP images with the previously generated FD image, and then compute the feature histogram. The indexing operation works as follows: in order to compute the histogram frequency (weight) of a given pixel value ($p_{x,y}^{R,N} | p_{x,y}^{R,N} \in [0, 2^{N-1}]$) of the LBP image $\text{LBP}_{x,y}^{R,N}$, we refer to the FD image $F_{x,y}(X \times Y)$. For all the pixel locations in the LBP image which have the same value as $p_{x,y}^{R,N}$, we find the sum of the FD values in all the same pixel locations in the FD image. This sum is the weight of the value $p_{x,y}^{R,N}$ in the histogram.

For instance, if the LBP value $p_{x,y}^{R,N}$ occurs at $n$ locations in the LBP image, namely $(x_1, y_1), (x_2, y_2), \ldots, (x_n, y_n)$, and the FD values of these $n$ locations in the FD image are $F(x_1, y_1), F(x_2, y_2), \ldots, F(x_n, y_n)$, then the weight $w$ of the LBP value in the feature histogram becomes

$$w = F(x_1, y_1) + F(x_2, y_2) + \ldots + F(x_n, y_n). \quad (16)$$

Mathematically, it can be expressed as:

$$\text{FWLBP}^{R,N}(p) = \mathcal{I}(\text{LBP}_{x,y}^{R,N}, F_{x,y}, p), \quad p \in [0, 2^{N-1}] \quad (17)$$

where

$$\mathcal{I}(\text{LBP}_{x,y}^{R,N}, F_{x,y}, p), = \begin{cases} \sum F_{x,y}, & \therefore \text{LBP}_{x,y}^{R,N} == p \\ 0, & otherwise \end{cases}$$

where $\text{FWLBP}(p)$ gives the weight of the value $p$ in the histogram, $\text{LBP}_{x,y}^{R,N}$ is the LBP image matrix constructed with sampling radius $R$ and $N$ sampling points, $F_{x,y}$ is the FD image as computed in Sec.( 2), and $\mathcal{I}$ is the indexing operation. The final FWLBP feature vector (Fig. 4(f)) is constructed by concatenating all $\text{FWLBP}(p)$ values. Since the weights of the LBP histogram are decided by the value of the fractal dimensions, we named the descriptor as *Fractal Weighted Local Binary Pattern*. Given a $m$ texture image, we have created the feature matrix $\mathcal{M}_{m \times n}$ where each row corresponds to the FWLBP descriptors of a texture. The principal component analysis (PCA) feature is computed as following. Initially, we compute the co-variance matrix $\Sigma$ of feature matrix as,

$$\Sigma(i,j) = \frac{\sum_{i=1}^{n}(\mathcal{M}(.,i) - \overline{\mathcal{M}(.,i)})(\mathcal{M}(.,i) - \overline{\mathcal{M}(.,i)})^T}{n-1} \quad (18)$$

where $\mathcal{M}_{m \times n}$ is the feature matrix, $n$ is the number of feature, $\mathcal{M}(.,i)$ represents the $i^{th}$ column of $\mathcal{M}$, and $\overline{\mathcal{M}(.,i)}$ is the mean of respective column. We calculate eigen-vector $\mathbf{e}$ of $\Sigma$ if it satisfies $\Sigma \mathbf{e} = \lambda \mathbf{e}$. Where $\lambda$ is an eigen-value of $\Sigma$. The eigen-values of $\Sigma$ are sorted in descending order as $\lambda_1 \geq \lambda_2 \geq \ldots \geq \lambda_n$ and corresponding eigenvectors $\mathbf{e}_1, \mathbf{e}_2, \ldots, \mathbf{e}_n$ are used as columns of linear transform matrix $\mathbb{U}$. Finally, $\mathcal{M}$ is multiplied by $\mathbb{U}^T$. Thus, the PCA features are provided by $\mathcal{D} = \mathbb{U}^T \mathcal{M}$. The scatter plot of proposed descriptor without and with PCA are shown in Fig. 5.

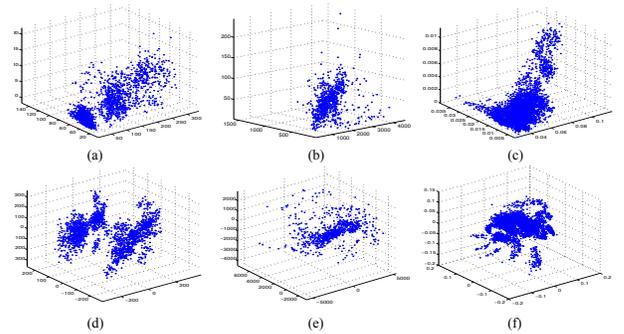

Figure 5: Top row (a)-(c) represent scatter plot of FWLBP feature, whereas bottom row (d)-(f) represent FWLBP feature after PCA transform, are extracted from Brodatz, KTH-TIPS and Outex_TC10 databases .

## 3.2 Complexity of FWLBP

**Algorithm 2:** Algorithm to extract $\text{FWLBP}_{R,N}^{x,y}$ descriptor, where $N$ is the number of circular samples taken at radius $R$.

**Data:** Input texture image, $A_{x,y}(X \times Y)$, FD image $F(x, y)$
**Result:** Normalized Fractal Weighted Local Binary Pattern (FWLBP) Descriptor

1. initialize ($R_{min} = 1, R_{max} = 3, N = 8, i = R+1, j = R+1, k = [0, K]$);
2. **while** ($R \leq R_{max}$) **do**
3.    /* Compute LBP image with sampling radius $R$, from a sample of $N$ local circularly spaced neighbours $P_{x,y}^{R,N}$ around a central pixel $A(x, y)$ */
4.    $\text{LBP}_{(x,y)R,N} = \sum_{n=1}^{N} 2^{(n-1)} \times sign(P_{x,y}^{R,N} - A(x, y))$;
5.    /* Compute weights of pixel values for histogram by indexing operation */
6.    $\text{FWLBP}(p) = I(\text{LBP}_{x,y}^{R,N}, F_{x,y}, p)$;
7.    $I(A, B, k) = \begin{cases} \sum B_{x,y}, & \text{for all } B_{x,y} \text{ where } A_{x,y} == k \\ 0, & \text{otherwise} \end{cases}$;
8.    /* Generate histrograms, concatenate, and normalize */
9.    $H_{R,N} = \text{HIST}(\text{FWLBP}_{R,N})$;
10.    $H_{\text{FWLBP}} = concatNormalize(\text{FWLBP}(H_{R,N}))$;
11. **end while**

Assuming the image is of size $M \times M$ and the kernel is of size $r \times r$, the algorithmic complexity of the proposed descriptors is computed as follows. To create the FD image of an input texture image with the number of blocks $\frac{M \times M}{r \times r}$ using Algorithm 1, the following operations are performed: $2M^2$ comparisons, $4(M/r)^2$ addition, $5(M/r)^2$ subtractions, $4(M/r)^2$ multiplication and $2(M/r)^2$ divisions. Therefore overall complexity of Algorithm 1 is $O(M^2)$. In Algorithm 2, assuming that neighboring pixels are considered with respect to each center pixel for the evaluation of the LBP for FD image of size $M \times M$. The time complexity to evaluate *Step* 4 of Algorithm 2 is $O(p \times M^2)$ [42] because $p$ circular shifting are required for each pixel, and there are $M^2$ pixels in total. *Step* 6 of Algorithm 2 takes $O(M^2)$ for Indexing operation and *Step* 9 takes $O(M^2)$ for building histogram. The overall time complexity to compute the proposed FWLBP descriptor using the combined Algorithms 1 and 2 is $O(M^2)$.

## 3.3 Comparing Distribution of FWLBP

After computing the FWLBP descriptors, as elaborated in the previous section, the distributions of FWLBP for the training (model) and test (sample) images are computed. The dissimilarity of model and test sample histograms can be measured using a non-parametric statistical test to find goodness-of-fit. Examples of metrics for evaluating the fit between two histograms are histogram intersection, log-likelihood ratio, and chi-square ($\chi^2$) statistic [40]. In this paper, the classification is performed via a *non-parametric* classifier called nearest subspace classifier (NSC) [43]. Here we have chosen NSC over popular *parametric* classifier like SVM due to the following reasons: the NSC finds an estimation of the underlying subspace within each class and assigns data points to the class that corresponds to its nearest subspace, which does not required any parameter tuning, while SVM has limitations in speed during both training and testing phase, and there is always a trade-off between the selection of kernel function, tuning of its hyper-parameters and classification performance. In order to avoid the over emphasizing patterns with large frequency, a preprocessing step is applied to the proposed feature before fit to NSC, similar to that in [44]:

$$\overline{X_k} = \sqrt{X_k},\ k = 1, 2, ..., N \quad (19)$$

where $N$ represents number of bins, and $X_k$ represents the original frequency of the LJP at $k^{th}$ bin. The nearest subspace classifier (NSC) first calculates the distance from the test sample y to the $c^{th}$ class and measures the projection residual $r_c$ from y to the orthogonal principle subspace $B_c \in \mathbb{R}^{N \times n}$ of the training sets $X_c$, which is spanned by the principal eigenvectors of $\sum_c = X_c X_c^T$ for the $c^{th}$ class, given as follows,

$$r_c = \|(I - P_{B_c})y\|_2 = \|(I - B_c B_c^T)y\|_2 \quad (20)$$

where $P = I \in \mathbb{R}^{N \times N}$ is a identity matrix where $N$ rows are selected uniformly at random. The test sample $y$ is then assigned to the one of the **C** classes with the smallest residual among all classes, i.e.

$$i^* = arg \min_{c=1,...,C} r_c \quad (21)$$

## 4 EXPERIMENTAL EVALUATION

### 4.1 Texture Databases

We have evaluated the performance of our proposed descriptor on five standard texture databases: Outex_TC-00010 (Outex_TC10) [5], Outex_TC-00012 (Outex_TC12) [5], Brodatz album [45], KTH-TIPS [46], and CUReT [47] texture database. We have compared the performance of the proposed FWLBP descriptor with $\text{LBP}_{R,N}$ [40], $\text{LBP}_{R,N}^{u2}$, $\text{DLBP}_{R,N}$ [42], $\text{LBP}_{R,N}^{sri\_su2}$ [48], multiscale $\text{CLBP}\_S_{R,N}^{riu2}/M_{R,N}^{riu2}/C(1, 8 + 3, 16 + 5, 24)$ [49], BIF [10] and other state-of-the-art descriptors. We have tested the classification accuracy using a non-parametric classifiers–NSC, and have reported the classification accuracy using $k$-fold cross-validation test. In $k$-fold cross-validation test, the feature set is randomly partitioned into $k$ equal sized subsets ($k = 10$). Out of the $k$ subsets, a single subset is retained as the validation data for testing the classifier, and the remaining $(k - 1)$ subsets are used as training data. The average of the classification accuracies over $k$ rounds give us a final cross-validation accuracy. We have normalized each input image to have an average intensity of 128 and a standard deviation of 20 [40]. In VZ-MR8 and VZ-Patch methods, the texture samples are normalized to have an average intensity of 0 and a standard deviation of 1 [11], [12], [18]. This is done to remove global intensity and contrast. The details of two experimental setups are given as follows:

Table 1: Summary of Texture Database used in Experiment #1

| Texture Database | Image Rotation | Illumination Variation | Scale Variation | Texture Classes | Sample Size (pixels) | Samples per Class | Total Samples |
|---|---|---|---|---|---|---|---|
| KTH-TIPS | ✓ | ✓ | ✓ | 10 | 200 x 200 | 81 | 810 |
| Brodatz | ✓ | | ✓ | 32 | 64 x 64 | 64 | 2048 |
| CUReT | ✓ | ✓ | | 61 | 200 x 200 | 92 | 5612 |

**EXPERIMENT #1:** **Brodatz** [45] album is chosen to allow a direct comparison with the state-of-the-art results [42]. There are 32 homogeneous texture classes Each image is partitioned into 25 non-overlapping sub-images of size

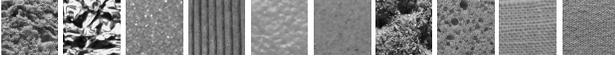

Figure 6: Ten texture images randomly taken from each class of KTH-TIPS database.

$128 \times 128$, and each sub-image is down-sampled to $64 \times 64$ pixels.

For the **CUReT** database [47], we have used the same subset of images as in [9], [11], [19], which contain 61 texture classes with 92 images per class. It is designed to contain large intra-class variation and is widely used to assess the classification performance. The images are captured under different illumination and viewing directions with constant scale. All 92 images of 61 texture classes are cropped into $200 \times 200$ region and converted to gray scale [11].

The **KTH-TIPS** database [46] is extended by imaging new samples of ten CUReT textures as shown in Fig. 6. It contains texture images with 3 different poses, 4 illuminations, and 9 different scales of size $200 \times 200$ and hence each class contains 81 samples. The **KTH-TIPS**, **Brodatz**, and **CUReT**, databases are summarized in Table 1.

Table 2: Summary of Texture Database used in Experiment #2

| Texture Database | Image Rotation | Illumination Variation | Scale Variation | Texture Classes | Sample Size (pixels) | Samples per Class | Total Samples |
|---|---|---|---|---|---|---|---|
| Outex_TC10 | ✓ | | | 24 | 128 x 128 | 180 | 4320 |
| Outex_TC12 | ✓ | ✓ | | 24 | 128 x 128 | 200 | 4800 |

**EXPERIMENT #2**: 24 different homogeneous texture classes are selected from the Outex texture databases [5], each having the size of $128 \times 128$ pixels. **Outex_TC_00010 (Outex_TC10)** contains texture with illuminant "inca", and **Outex_TC_00012 (Outex_TC12)** contains textures with illuminants "inca", "horizon", and "tl84". Both of the Outex test suites are collected under 9 different rotation angles $(0°, 5°, 10°, 15°, 30°, 45°, 60°, 75°,$ and $90°)$ in each texture class. The test suites **Outex_TC10**, and **Outex_TC12** are summarized in Table 2. Running experiments on the whole database is challenging due to a large number of texture classes, small number of samples per class, and lack of intra-class variation.

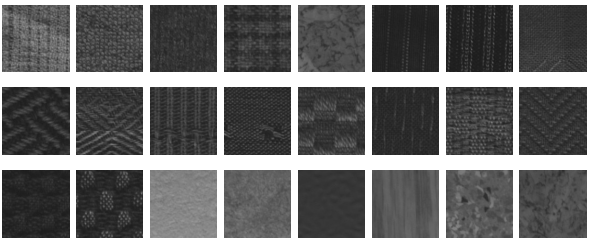

Figure 7: 24 texture images randomly taken from each class of Outex_TC10 and Outex_TC12 database

### 4.2 Results of Experiment #1

To test the scale invariance performance of the proposed descriptor, we have used the KTH-TIPS database, which contains relatively small variations in scales (i.e, 9 scales, continually from 0.5 to 2) [46]. In literature, for texture classification task usually LBP variants were used, in this section, we compared FWLBP descriptor with some powerful variant of LBPs, which includes CLBP [49], LBPV [50], DLBP [42], LBP$_{R,N}^{sri\_su2}$ [48], PRICOLBP [52], BRINT [51], and COALBP [55]. The classification performance is evaluated on three well known benchmark texture databases (KTH-TIPS, Brodatz, and CUReT) and results are shown in Table 3. The observations noted from Table 3 are as follows: The scale-invariant LBP$_{R,N}^{sri\_su2}$ descriptor performs better than LBP$_{R,N}^{riu2}$. However, the performance is worse than multi-resolution LBP$_{R,N}^{riu2}$ descriptor, and CLBP_$S_{R,N}^{riu2}/M_{R,N}^{riu2}/C$ descriptor; and much worse than the proposed descriptor. This is because extracting consistent and accurate scale for each pixel is difficult. However, the LBP$_{R(i,j),8}^{sri\_su2}$ provides good performance in controlled environment [48], but it fails over more complex databases. The DLBP, when combined with Gabor features, attains a higher classification rate than the conventional LBP with NNC. However, its performance is quite less than the proposed FWLBP, as it does not consider changes in scale. FWLBP achieves remarkably better classification performance than CLBP, LBPV, LBPHF_$S$ and LBPHF_$S\_M$ on KTH-TIPS, Brodatz and CUReT, texture datasets and yields comparable performance with DLBP. Note that LBP, LBPHF$_S$, LBPHF_$S\_M$, CLBP, and LBPV generally encode multi-scale information whereas the proposed FWLBP descriptor encodes the spatial fractal dimension to characterize the local scale information into the 1-$\mathbb{D}$ LBP histogram.

Table 3: Comparison with other Variants of LBP

| Methods | Classifier | Classification Accuracy (%) | | |
|---|---|---|---|---|
| | | KTH-TIPS [46] | Brodatz [45] | CUReT [47] |
| LBPV [50] | NNC | 95.50 | 93.80 | 94.00 |
| BRINT [51] | NNC | 97.75 | 99.22 | 97.06 |
| LBP$_{8,8}^{riu2}$ [40] | NNC | 82.67 | 82.16 | 80.63 |
| DLBP$_{3,24}$ [42] | SVM | 86.99 | 99.16 | 84.93 |
| LBP$_{1,8}^{sri\_su2}$ [48] | NNC | 89.73 | 69.50 | 85.00 |
| LBP$_{(1,8+2,16+3,24)}$ [40] | NNC | 95.17 | 91.60 | 95.84 |
| CLBP_SMC [49] | NNC | 97.19 | 94.80 | 97.40 |
| SSLBP [44] | NNC | 97.80 | - | 98.55 |
| PRICOLBP$_g$ [52] | SVM | 98.40 | 96.90 | 98.40 |
| LBPHF_S [53] | NNC | 97.00 | 94.60 | 95.90 |
| LBPHF_S_M [54] | NNC | 97.00 | 94.60 | 95.90 |
| COALBP [55] | NNC | 97.00 | 94.20 | 98.00 |
| LMP [25] | NNC | 98.37 | - | 98.11 |
| Proposed FWLBP | NSC | 99.75 | 99.62 | 99.10 |

Table 4: Texture Classification Results on KTH-TIPS, Brodatz, and CUReT

| Methods | Classifier | Classification Accuracy (%) | | |
|---|---|---|---|---|
| | | KTH-TIPS [46] | Brodatz [45] | CUReT [47] |
| VZ-MR8 [11] | NNC | 94.50 | 94.62 | 97.43 |
| VZ-Patch [12] | NNC | 92.40 | 87.10 | 98.03 |
| Lazebnik et al. [21] | NNC | 91.30 | 88.20 | 72.50 |
| Zhang et al. [9] | SVM | 96.10 | 95.90 | 95.30 |
| Liu et al. [19] | SVM | - | 94.20 | 98.50 |
| MFS [26] | NNC | 81.62 | - | - |
| Capato et al. [56] | SVM | 94.80 | 95.00 | 98.50 |
| PFS [6] | SVM | 97.35 | - | - |
| BIF [10] | Shift NNC | 98.50 | 98.47 | 98.60 |
| Proposed FWLBP | NSC | 99.75 | 99.62 | 99.10 |

Table 4 shows performance of texture classification performance other than LBP variants. Lazebinik *et al.* [21] proposed to detect interest regions using Harris-affine corner and Laplacian-affine blobs and then extracted SPIN and RIFT as texture signatures after normalizing these regions. Finally, texture classification is performed using nearest neighbor classifier (NNC). Caputo et al. [56] used SVM kernel instead of NNC and reveal that the SVM classifier could achieve reasonably better performance. Zhang et al. [9] proposed object and texture classification by analyzing different texture features and kernels. Recently, global scale invariant feature extraction methods drew attention because local scale normalization is usually slow due to pixel by pixel operations. Xu et al. [22] and Quan et al. [6] have classified the image pixels into multiple point sets by gray intensities or local feature descriptors. Multi-scale BIF [10] at scales $\sigma, 2\sigma, 4\sigma$, and $8\sigma$ gives better performance than the other bag-of-words methods when NNC classifier is used. This is mainly because BIF uses pyramid histogram with time inefficient shift matching scheme. Also, the feature dimension of BIF descriptor [10], CLBP [49], SSLBP [44] and PRI-COLBP [52] are larger ($6^4 = 1296, 2200, 480 \times 5 = 1000$ and $590 \times 2 = 1180$) than the dimension of proposed FWLBP descriptor ($256 \times 3 = 768$). The performance of BIF is reduced when scale shifting scheme is not considered [10]. Apart from bag-of-words model, the variant of LBP based descriptors, e.g., CLBP *, LBPV, LTP, DLBP, $LBP_{R,N}^{sri\_su2}$, PRICOLBP, BRINT, COALBP, and SSLBP also perform reasonably well on texture classification task. The observations noted from Table 4 are as follows: FWLBP also exceeds the classification performance of several bag-of-words methods, [9], [12], [19], [21], [55], [56]. Fig. 9 shows the graph of success rates against the number of selected feature on Brodatz, KTH-TIPS and Outex_TC10 texture databases. The classification rate of proposed descriptor rapidly increases due to PCA transform, which carries the most relevant information with large variance among the first component of the feature matrix. It is interesting to note that even for the database having large number of variations and a few number of samples, this property can make the proposed descriptor suitable for other applications where the number of features plays an important role. In this work the dimension of the proposed FWLBP reduces to 300 using PCA.

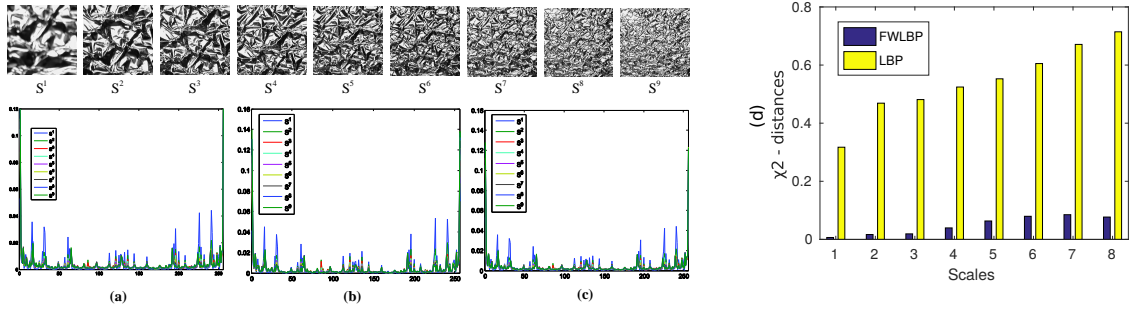

Figure 8: (a)-(c) represent the individual histograms ($H_{R,N}$ of FWLBP, where $R = 1, 2, 3$ and $N = 8$) for a texture sample taken from the KTH-TIPS database [46], having 9 different scales. The abscissa and ordinate represent number of bins and feature probability distribution, respectively. As the histograms of individual FWLBP for the 9 different scales are overlapping, it implies that the FWLBP descriptor is scale invariant. (d) represents the Chi-square ($\chi^2$) distance of proposed FWLBP and LBP descriptor extracted from original texture image ($S^1$) and 8 texture image with different scale ($S^2 - S^9$).

Although the images are captured under scale, rotation, and illumination variations, the proposed FWLBP gives sound performance as given by results in Table 4, and achieves comparable, and sometimes even better perfor-

*We used the best CLBP settings i.e., CLBP_S/M/C in this paper mance, than the state-of-the-art methods. It validates that the encoding strategy preserves the relative information about local scale change along with orientation angle are effective. An illustration of the scale invariance property of the proposed descriptor is given in Fig. 8, where $S^1$-$S^9$ show nine KTH-TIPS texture images of "corduroy" class at different scales from 0.5 to 2, and (a)-(c) depict the histograms ($H_{R,N}|R \in [1, 2, 3]$) of individual $FWLBP_{R,N}$ descriptor with $R = 1, 2, 3$, and $N = 8$. It is evident from Fig. 8(a)-(c) that the histograms of individual FWLBP for the 9 different scales are approximately identical. In addition Fig. 8(d) shows the Chi-square ($\chi$) distance [12] between extracted FWLBP and LBP descriptor from original texture image ($S^1$) and 8 texture images with different scale ($S^2 - S^9$), its shown that proposed descriptor has less $\chi$-distance compared to LBP. Therefore, from both the observations it implies that the FWLBP descriptor is not sensitive to small scale variations. The proposed descriptor achieves scale invariance by utilizing the properties of Fractal Dimension and Gaussian Scale Space, as explained in Theorem 1. Fig. 10(a)-(b) represent images of two texture classes taken form Brodatz database and Fig. 10(c) shows PCA transformed FWLBP descriptor extracted from three images per texture class and respective feature probability distribution functions (PDF). Since both texture classes contain very small intra-class variation even though their PDF provides a strong indication about the discriminative power of the proposed descriptor.

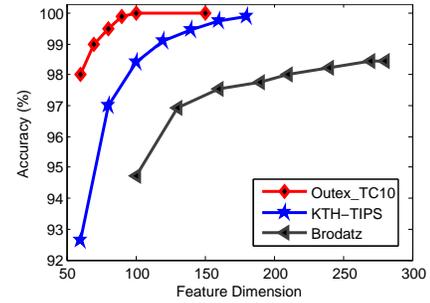

Figure 9: Success rate against the number of descriptors on Brodatz, KTH-TIPS and Outex_TC10 texture databases

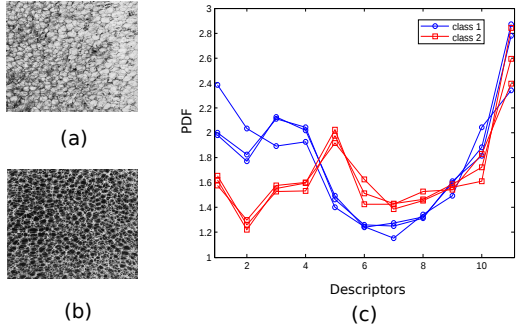

Figure 10: Discriminative power of the proposed FWLBP descriptors are visualized using PCA. We have taken 6 images from two classes of Brodatz textures and extracted the FWLBP. Notice that the classes are substantially separated by the curves.

### 4.3 Analysis of parameter $r_{max}$

The proposed fractal based descriptor has only one hyper-parameter $r_{max}$, which is the maximum value of the scaling factor of the DBC algorithm. The $r_{max}$ represents how much a specific intrinsic structure is *self-similar* [38] to its surrounding pixels. The different values of $r_{max}$ influence the deviation of classification accuracies. Table 5 shows recognition accuracy for Brodatz textures using proposed FWLBP and FWLBP$^{u2\dagger}$, where $r_{max}$ values are ranging from 2 to 7. Its shows clearly that both the descriptors have similar trends towards $r_{max}$ and recognition rate increases till it reaches $r_{max}$ to 7. Table 5 also indicates that for different scaling factors the proposed FWLBP descriptor significantly outperforms FWLBP$^{u2}$ on well-known Brodatz texture database (includes images of scale and rotation or view-point variations). To compute fractal dimension of an image of size less than let say, $200 \times 200$ fast, Faraji and Qi [41] recommend $r_{max}$ to be used in between 7 and 10 to compromise between good texture recognition accuracy and computational overheads. In our experimental setup, we use 7 as $r_{max}$ value for all the texture databases.

Table 5: The Classification accuracies of proposed FWLBP and FWLBP$^{u2}$ descriptors using different scaling factors of DBC algorithm on brodatz texture database.

| Methods | Classification Rates With Different box sizes ranges $r_{min}$ to $r_{max}$ | | | | | |
|---|---|---|---|---|---|---|
| | $2 \times 2$ | $2 \times 2$ to $3 \times 3$ | $2 \times 2$ to $4 \times 4$ | $2 \times 2$ to $5 \times 5$ | $2 \times 2$ to $6 \times 6$ | $2 \times 2$ to $7 \times 7$ |
| Proposed FWLBP | 99.24 | 99.25 | 99.27 | 99.27 | 99.31 | 99.62 |
| Std. | ±0.9216 | ±0.9688 | ±0.7547 | ±0.6701 | ±0.6484 | ±0.3257 |
| Error | 0.0052 | 0.0052 | 0.0052 | 0.0104 | 0.0104 | 0.0002 |
| Proposed FWLBP$^{u2}$ | 98.05 | 98.16 | 98.30 | 98.45 | 98.53 | 98.57 |
| Std. | ±0.6717 | ±0.8426 | ±0.8590 | ±0.8058 | ±0.7514 | ±0.7243 |
| Error | 0.0156 | 0.0208 | 0.0208 | 0.0156 | 0.0104 | 0.0260 |

### 4.4 Results of Experiment #2

The results of Experiment #2, carried out on Outex_TC10 and Outex_TC12 are tabulated in Table 6. It shows the average classification accuracy obtained by $k$-fold cross-validation test ($k = 10$) for proposed descriptor, and the comparative summary of the results for variants of LBP e.g.,

$\dagger$Where 'u2' indicates *uniform pattern* encoding;

LBP$_{(R,N)}$ [40], VZ-MR8 [11], VZ-Patch [12] and recent state-of-the-art bag-of-word methods. The observations made from the results of Experiment #2 are stated as follows. LBP and LBPV have similar feature dimensions, but the later one incorporates additional contrast measures to the pattern histogram, it produces a significant performance improvement compared to the conventional LBP. LBP$_{R,N}^{riu2}$/VAR$_{R,N}$ provides better performance compared to LBPV$_{R,N}^{riu2}$, because LBP and local variance of a texture image are complementary; and hence the joint distribution of LBP and local variance gives better results than any one alone. CLBP_$S/M/C$ is created by fusing CLBP_$S$ and CLBP_$M/C$. It provides better performance compared to other variants of CLBP as it has complementary features of sign and magnitude, in addition to the center pixel representing the gray value of the local patch. CDCP [24] achieves improved performance over CLBP since in CDCP patterns are extracted from component and holistic levels. The state-of-the-art bag-of-words statisti-cal algorithms, VZ-MR8 and VZ-Patch, take dense response from multiple filters. However, their performance is poor compared to the proposed FWLBP. Also, feature extraction and matching complexity of these two techniques is quite high [12], when compared to the proposed FWLBP due to the MR8 needs to find 8 maximum responses after 38 filters convolving with the image and compares every 8-dimension vector in an image with all the textons to build histograms using clustering technique. DLBP + NGF utilizes the top ranking 80% of LBP pattern to improve the recognition, as compared with the results obtained using original LBP$_{R,N}^{u2}$. But like VAR$_{R,N}$, it neglects local spatial structure which is important for texture discrimination. Also, DLBP needs pre-training stage, and the dimensionality of the DBLP varies with the training samples. For comparison, we have taken the best results of DLBP with $R = 3$ and $N = 24$ in Table. 6. The LBP$_{R,N}^{NT}$ [59] based methods and BRINT [51] give better performance compared to other state-of-the-art LBP methods. However, compared to the proposed FWLBP, their accuracies are lower. This is because LBP$_{R,N}^{NT}$ extracts features by locally rotation invariant LBP$_{R,N}^{riu2}$ approach, which produces only 10 bins. Such small features cannot represent each class well. On the other hand, BRINT extracts a large number of features from multiple resolution (R = 1, 2, 3, 4) by utilizing rotation invariant LBP$_{R,N}^{ri}$ approach, but it loses some global image information. From Table 6, it is evident that our proposed descriptor provides better classification performance compared to other state-of-the-art methods. The better performance is due to generation of a six-layer Gaussian Scale Space to provide basic scale-invariance (Fig. 1), transforming the scale space to fractal dimension (FD) images, which also provides additional levels of scale invariance and some degree of illumination invariance (Sec. 3), and finally using the FD as weights for the values in the feature vector histograms. When all these are put together we achieve high degree of invariance to scale, rotation, and reflection, and are able to capture the *micro structure* in the different illumination controlled environments (for images like "inca", "horizon" and "tl84"). To visualize the rotation invariant characteristic of our proposed descriptor, an example of the FWLBP feature distribution for a texture sample is taken from the Outex_TC10 database (with 9 orientations – $0°, 5°, 10°, 15°, 30°, 45°, 60°, 75°$, and $90°$) and is shown in Fig. 11. Fig 11(a)-(c) represent the histograms ($H_{R,N}|R \in [1,2,3]$) of individual ($FWLBP_{R,N}|R \in [1,2,3]$). The histograms show that the FWLBP feature distribution for the different texture orientations are approximately overlapping, which signify the rotation invariance property of the FWLBP descriptor. In addition, Fig. 11(c) represents the histograms of individual FWLBP for the 9 different rotational angles, which are approximately identical. In addition, Fig. 8(d) shows the Chi-square ($\chi$) distance [12] between extracted FWLBP and LBP features from original texture image ($0°$) and 8 texture images with different rotational viewpoints ($5° - 90°$), it is shown that the proposed descriptor has less $\chi$-distance compared to LBP.

Table 6: Average classification accuracy (%) on Outex_TC10 and Outex_TC12 using state-of-the-art schemes

| Method | Classifier | Outex_TC10 | Outex_TC12 | | Average |
|---|---|---|---|---|---|
| | | | horizon | tl84 | |
| VZ-MR8 [11] | NNC | 93.59 | 92.82 | 92.55 | 92.99 |
| VZ-Patch [12] | NNC | 92.00 | 92.06 | 91.41 | 91.82 |
| LTP [57] | NNC | 76.06 | 63.42 | 62.56 | 67.34 |
| VAR [58] | NNC | 90.00 | 64.35 | 62.93 | 72.42 |
| LBP [40] | SVM | 97.60 | 85.30 | 91.30 | 91.40 |
| LBP$_{R,N}^{riu2}$ | NNC | 84.89 | 63.75 | 65.30 | 71.31 |
| LBP/VAR | NNC | 96.56 | 78.08 | 79.31 | 84.65 |
| LBPV$_{R,N}^{riu2}$ [50] | NNC | 91.56 | 77.01 | 76.62 | 81.73 |
| CLBP_$S/M/C$ [49] | NNC | 98.93 | 92.29 | 90.30 | 93.84 |
| LBP$_{R,N}^{NT}$ [59] | NNC | 99.24 | 96.18 | 94.28 | 96.56 |
| DLBP$_{R=3,N=24}$ [42] | SVM | 98.10 | 87.40 | 91.60 | 92.36 |
| BRINT_Cs_CM [51] | NNC | 99.35 | 97.69 | 98.56 | 98.12 |
| PTP [60] | NNC | 99.56 | 98.08 | 97.94 | 98.52 |
| CDCP [24] | NSC | 99.76 | 99.82 | 99.62 | 99.72 |
| LMP [25] | NNC | 99.88 | 99.79 | 99.76 | 99.81 |
| Proposed FWLBP | NSC | 99.97 | 99.98 | 99.93 | 99.96 |

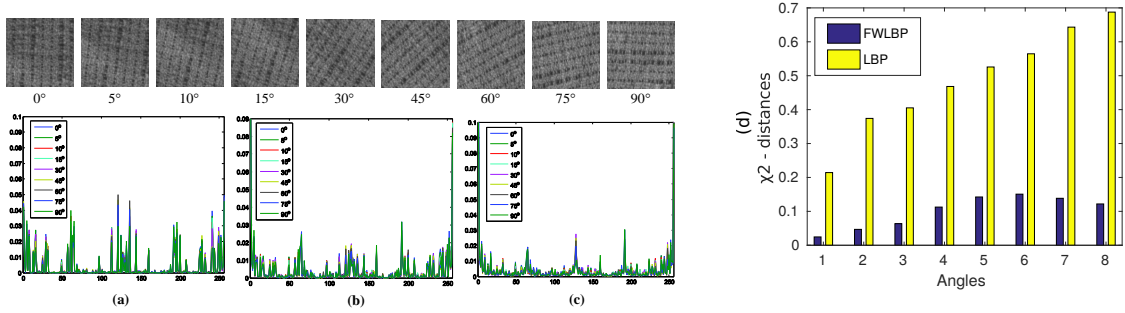

Figure 11: (a)-(c) represent individual FWLBP histograms ($FWLBP_{R,N}|R \in [1,2,3]$) of a texture sample taken from the Outex database, with 9 different orientations. The abscissa and ordinate represent the number of bins and feature probability distribution, respectively. The histograms show that the feature distribution for the different texture orientations are approximately overlapping, which signify the rotation invariance property of the proposed FWLBP descriptor. (d) represents the Chi-square ($\chi$) distance between original texture image ($0°$) and 8 texture image with different angles ($5° - 90°$)

Though the trend is clear from the performance Table 6, we have further analysed the performance using one way statistical analysis of variance (ANOVA) test [61]. ANOVA is a collection of statistical tests used to analyze the differences among group means and their associated procedures. In other word, one-way ANOVA is used to test the equality of two or more means at one time using the variances. The null hypothesis $H_0$ for the test is, *there is no significant difference among group means*. We have taken the significance level $\alpha = 0.05$ for this ANOVA test. We can reject $H_0$ if the $p$-value for an experiment is less than the selected significant level, which implies that the at least one group mean is significantly different from the others. To understand why the performance of proposed FWLBP descriptor was significantly different from well-known descriptors such as VZ-Patch, VZ-MR8, BRINT, DLBP, CLBP, and LBP$^{riu2}$, we conduct one way ANOVA test with significance level $\alpha = 0.05$. The test results are shown in Table 7 where the $p$-value ($1.6460\mathrm{e}^{-07}$) is much lesser than the pre-select significance level $\alpha = 0.05$. This indicates that the performance of proposed descriptor significantly better than other descriptors and hence we can reject the hypothesis $H_0$. In addition, the box plot corresponding to aforementioned ANOVA test is shown in Fig. 13, which also clearly indicates that the mean performance of FWLBP descriptor is significantly better than the well-known descriptors such as VZ-MR8 [11], VZ-Patch [12], BRINT [51], DLBP [42], CLBP [49], and LBP$^{riu2}$ [40].

Table 7: One way statistical ANOVA test results for Outex, KTH-TIPS, Brodatz, and CUReT databases, where the significance level $\alpha$ is 0.05.

| Source | SS | df | MS | F | Prob (p) > F |
|---|---|---|---|---|---|
| Groups | 1923.34 | 06 | 325.557 | 14.36 | $1.6460\mathrm{e}^{-07}$ |
| Error | 0621.82 | 28 | 022.208 | | |
| Total | 2515.67 | 34 | | | |

The performances of the proposed FWLBP are also compared with variants of LBP using Cumulative Match Characteristic (CMC) curve [62] and the comparative results are shown in Fig. 12. The CMC curve also show that the proposed FWLBP provides better performance compared to other state-of-the-art LBP methods.

### 4.5 Robustness to Noise

To evaluate the performance of the proposed method in noisy environment, the experiments are carried out on Outex_TC10 texture database (discussed in subsec. 4.1) by adding white Gaussian noise, resulting in different Signal to Noise Ratio (SNR). The training and testing set-up are the same as in noise-free situation.

Table 8 demonstrates the noise robustness of different methods on Outex_TC10 database based on classification rates of various noise levels (measured using SNR i.e Signal to Noise Ratio). The proposed descriptor achieves state-of-the-art results in term of average accuracy of 100%, 100%, 99.97%, 99.95%, and 99.93% on SNR = 100 dB, 30 dB, 15 dB, 10 dB, and 5 dB, respectively. It can be observed from Table 8 that the proposed FWLBP descriptor is very modestly improved as compared with other LBP based descriptors. The proposed descriptor has inherited this property from the calculation of fractal dimension in Gaussian scale space representation in contrast to the LBP and its variants.

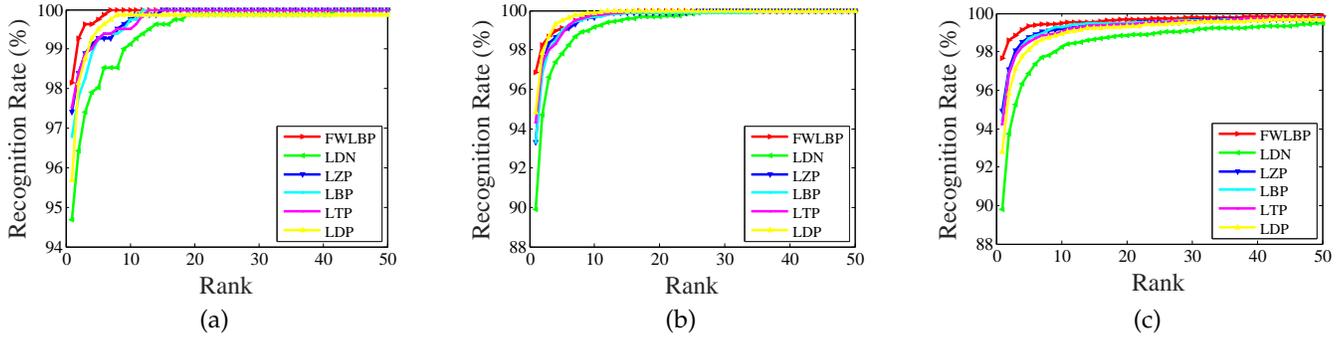

Figure 12: Cumulative Match Characteristic (CMC) curve of Variants of LBP for (a) KTH-TIPS (b) Brodatz and (c) CUReT texture dabases

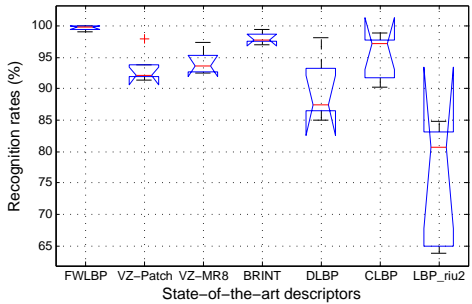

Figure 13: The box plot (Descriptor vs. Accuracy) corresponding to one way statistical ANOVA test for proposed FWLBP and state-of-the-art descriptors on Outex, KTH-TIPS, Brodatz, CUReT databases.

Table 8: Classification Accuracy (%) of Proposed Method and Different state-of-the-art Methods on Outex_TC10 with Different Noise Levels in term of Db.

| Methods | Classifier | Classification Accuracy (%) | | | | |
|---|---|---|---|---|---|---|
| | | SNR = 100 | SNR = 30 | SNR = 15 | SNR = 10 | SNR = 5 |
| $LBP_{R,N}^{riu2}$ [40] | NNC | 95.03 | 86.93 | 67.24 | 49.79 | 24.06 |
| $LBP_{R,N,k}^{NT}$ [59] | NNC | - | 99.79 | 99.76 | 99.76 | 99.74 |
| CLBP_SMC [49] | NNC | 99.30 | 98.12 | 94.58 | 86.07 | 51.22 |
| $LTP_{R=3,N=24}^{riu2}$ [57] | NNC | 99.45 | 98.31 | 93.44 | 84.32 | 57.37 |
| $NRLBP_{R,N}^{riu2}$ [63] | NNC | 84.49 | 81.16 | 77.52 | 70.16 | 50.88 |
| BRINT [51] | NNC | 97.76 | 96.48 | 95.47 | 92.97 | 88.31 |
| Proposed FWLBP | NSC | 100 | 100 | 99.97 | 99.95 | 99.93 |

### 4.6 Comparison of Proposed FWLBP with state-of-art deep learning based methods

A key characteristic of deep convolutional neural networks (CNN)(see Fig. 14) is a hierarchical representation which is universal and directly generated from the data, used to perform image classification task. Deep CNNs have universal shown their power of pattern recognition. However, the robustness for recognition is limited due to the lack of geometric invariance of global CNN activations. An effective texture descriptor FV-CNN has been introduced by Cimpoi *et al.* [8], where at first CNN features are extracted at multiple scales. Then an order-less Fisher Vector pooling operation is performed. Despite significant progress of deep CNN models, not much analytical insights into its internal operation and behaviour is available. Mallat *et al.* [64], [65] have introduced scale and rotation invariant wavelet convolution scattering network (ScatNet) where the convolution filters are pre-defined as wavelet and no learning process is needed. Inspired by ScatNet, a simple deep learning network, PCANET is proposed by chan *et al.* [66] which is based on cascading of multistage PCA, binary hashing and histogram pooling.

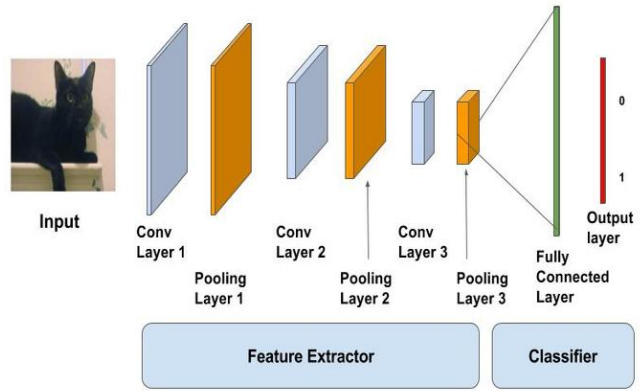

Figure 14: A Deep CNN architecture where the input is processed in a feed forward manner through the stage of convolutions and sub-sampling and finally classified with a linear or non-linear classifier.

A simple variation of PCANET, named RANDNET, in which the cascaded filters are randomly selected but not learned, was also proposed by Chan *et al.* [66]. One of the major trends in deep CNN research community is to use more and more complex networks to improve the classification performances. However, it needs the powerful and large memory computer and GPUs to train very deep and computationally expensive networks. The comparative results of texture classification performance with feature dimensionality of the proposed FWLBP and state-of-art deep learning based methods such as FV-CNN [8], [67], SCATNET [64], [65], PCANET, PCANET$^{riu2}$, and RANDNET [66], are tabulated in Table 9. It is observed from Table 9 that the proposed FWLBP provides comparable or better classification performance compared to state-of-art deep learning based methods. In addition, the feature dimension of proposed FWLBP descriptor is less than the deep learning based mod-

els[‡].

Table 9: The texture classification performance and feature dimensionality of the proposed FWLBP and state-of-art deep learning based methods

| Methods | Classification Accuracy (%) | | | | Feature Dimention |
|---|---|---|---|---|---|
| | Outex_TC10 | Outex_TC12 | KTH-TIPS | CUReT | |
| FV-VGGDM (SVM) [8], [67] | 80.00 | 82.30 | **88.20** | 99.00 | 65536 |
| FV-AlexNet (SVM) [67], [69] | 67.30 | 72.30 | 77.90 | 98.40 | 32768 |
| SCATNET (PCA) [64], [65] | 99.69 | 99.06 | 69.92 | **99.66** | 596 |
| SCATNET (NNC) [64], [65] | 98.59 | 98.10 | 63.66 | 95.51 | 596 |
| PCANET (NNC) [66] | 39.87 | 45.53 | 59.43 | 57.70 | 2048 |
| PCANET$^{riu2}$ (NNC) [66] | 35.36 | 40.88 | 52.15 | 81.48 | **80** |
| RANDNET (NNC) [66] | 47.43 | 52.45 | 60.67 | 90.87 | 2048 |
| RANDNET$^{riu2}$ (NNC) [66] | 43.54 | 45.70 | 56.90 | 80.46 | **80** |
| Proposed FWLBP (NSC) | **99.97** | **99.96** | **99.75** | 99.10 | 300 |

## 5 CONCLUSIONS

In this paper, we have proposed a simple, efficient yet robust descriptor, called the *Fractal Weighted Local Binary Pattern* (FWLBP) for texture classification. At first, a Gaussian Scale Space representation of the texture image is generated, and then Differential Box Counting (DBC) algorithm is used to transform the images into fractal dimension. An augmented form of Local Binary Pattern (LBP), with fixed sample size and varying radius is used, and the weights for the samples are calculated using an indexing function. Finally, the normalized feature vector is formed by concatenating the histogram of FWLBP for all elements of LBP. Experimental results show that our proposed FWLBP descriptor provides promising performance under scale, rotation, reflection, and illumination variation. Comparative study indicates that our descriptor also performs better, as compared to other state-of-the-art methods. In addition, the dimensionality of the extracted feature vector is significantly less the state-of-the-art method, making it computationally efficient. As it offers reasonably high speed to extract the features, this descriptor may be applied to real time scenarios.